\documentclass[conference, compsoc]{IEEEtran}
\usepackage[backend=bibtex,bibstyle=ieee,citestyle=numeric-comp]{biblatex}
\AtBeginBibliography{\small}
\bibliography{bibliography}
\IEEEoverridecommandlockouts
\usepackage{amsmath,amssymb,amsfonts}
\usepackage{algorithm}
\usepackage{algpseudocode}
\usepackage{graphicx}
\usepackage{subcaption}
\usepackage{textcomp}
\usepackage{xcolor}
\usepackage{mathtools}
\usepackage{dsfont}
\usepackage{tabularx}
\usepackage{todonotes}
\usepackage{verbatim}
\newcolumntype{Y}{>{\centering\arraybackslash}X}
\usepackage{caption} 
\usepackage[normalem]{ulem}
\usepackage[noabbrev]{cleveref}

\newcommand{\argmin}[1]{\underset{#1}{\operatorname{argmin}}\;}

\newcommand{\linebreakand}{%
  \end{@IEEEauthorhalign}
  \hfill\mbox{}\par
  \mbox{}\hfill\begin{@IEEEauthorhalign}
}

\title{ResBuilder: Automated Learning of Depth with Residual Structures}

\author{
\IEEEauthorblockN{Julian Burghoff\IEEEauthorrefmark{1},
Matthias Rottmann\IEEEauthorrefmark{2},
Jill von Conta\IEEEauthorrefmark{3},\\
Sebastian Schoenen\IEEEauthorrefmark{4},
Andreas Witte\IEEEauthorrefmark{5},
Hanno Gottschalk\IEEEauthorrefmark{6}}
\IEEEauthorblockA{\IEEEauthorrefmark{1}\textit{Department of Mathematics}, \textit{University of Wuppertal}, Wuppertal, Germany, burghoff@math.uni-wuppertal.de}

\IEEEauthorblockA{\IEEEauthorrefmark{2}\textit{Department of Mathematics}, \textit{University of Wuppertal}, Wuppertal, Germany, rottmann@uni-wuppertal.de}

\IEEEauthorblockA{\IEEEauthorrefmark{3}\textit{Research \& Development}, \textit{Control Expert}, Langenfeld, Germany, j.vonconta@controlexpert.com}

\IEEEauthorblockA{\IEEEauthorrefmark{4}\textit{Research \& Development}, \textit{Control Expert}, Langenfeld, Germany,  s.schoenen@controlexpert.com}

\IEEEauthorblockA{\IEEEauthorrefmark{5}\textit{Control Expert}, Langenfeld, Germany, a.witte@controlexpert.com}

\IEEEauthorblockA{\IEEEauthorrefmark{6}\textit{Institute of Mathematics}, \textit{TU-Berlin}, Berlin, Germany, gottschalk@math.tu-berlin.de}
}

\begin{document}

\maketitle

\begin{abstract}
In this work, we develop a neural architecture search algorithm, termed Resbuilder, that develops ResNet architectures from scratch that achieve high accuracy at moderate computational cost. It can also be used to modify existing architectures and has the capability to remove and insert ResNet blocks, in this way searching for suitable architectures in the space of ResNet architectures. In our experiments on different image classification datasets, Resbuilder achieves close to state-of-the-art performance while saving computational cost compared to off-the-shelf ResNets. Noteworthy, we once tune the parameters on CIFAR10 which yields a suitable default choice for all other datasets. We demonstrate that this property generalizes even to industrial applications by applying our method with default parameters on a proprietary fraud detection dataset.
\end{abstract}

\begin{IEEEkeywords}
automated machine learning, neural architecture search, residual neural networks
\end{IEEEkeywords}

\section{Introduction}
Machine learning is a key technology for data-driven automation that achieved intriguing success in many applications, such as image recognition \cite{russakovsky2015imagenet} and natural language processing \cite{nakano2021webgpt}.
However, the application of machine learning to ever new tasks requires data scientists with the skill to design problem specific machine learning algorithms. In particular this applies when working with deep neural networks \cite{goodfellow2016deep} and the question arises how to choose network architectures and select the hyper parameters to control the training process. Especially for small and medium size enterprises, this constitutes a major obstacle in the application of machine learning in their processes or products. Automated Machine Learning (AutoML) \cite{he2021automl, choudhary2020comprehensive} is a research area that studies hyperparameter optimization \cite{yao2018taking} as well as neural architecture search (NAS) \cite{elsken2019neural}.

Many of the existing methods in the field of network architecture search (NAS) modify existing network architectures \cite{he2021automl}. In many cases, runtime optimization of a baseline neural network while preserving accuracy of predictions is a motivation for many of the works in this field \cite{hsu2018monas}.

Another field closely related to NAS is the task of network pruning \cite{reiners2022efficient, he2017channel} where an existing network is made sparser in order to reduce the computational burden. More drastically, complete layers are removed in \cite{srinivas2015data, liu2018rethinking}.

In this work, we introduce ResBuilder that tackles the problem of constructing ResNet \cite{he2016deep} architectures from scratch. The goal is to find a ResNet architecture in an automated way that achieves high accuracy for a given problem while not exceeding a predefined computational budget. 
More precisely, we utilize MorphNet \cite{Gordon17} as a baseline which performs channel pruning as well as layer removal during training. We introduce a method to add and remove layers during training, dynamically controlling the network's capacity while balancing test accuracy and computational expense.

To achieve this, we focus on ResNet blocks that support layer insertion and removal during training in a natural way. 
Due to the skip connection, layers with weights close to zero almost act as identity layers. Such layers can thus be inserted during training without undoing the previous training progress. Similarly, layers with weights small in magnitude can be seamlessly removed during training with undoing previous progress. During architecture search, we utilize a layer LASSO approach in order to identify unnecessary layers of the parameters.

We demonstrate the efficiency of ResBuilder on six datasets, namely Animals10 \cite{animals10dataset}, CIFAR10 \cite{CIFAR10}, CIFAR100 \cite{CIFAR10}, MNIST \cite{lecun1998MNIST}, FashionMNIST \cite{FashionMNIST}, EMNIST \cite{cohen2017emnist}, and study its hyperparameters in-depth. 
It turns out that ResBuilder easily builds ResNet architectures achieving close to state-of-the-art performance (without pre-training) on a variety of image classification benchmarks while saving computational cost compared to off-the-shelf ResNets.

Our contribution can be summarized as follows:

\begin{itemize}
    \item We develop ResBuilder which constructs ResNet architectures from scratch and modifies existing ones to achieve high accuracy at a given computational budget.
    \item We achieve close to state-of-the-art accuracy on various academic benchmark data sets with a default hyperparameter setting that was tuned once for CIFAR10.
    \item We showcase the effectiveness of ResBuilder for the industrial application of detecting manipulated images. Specifically, we focus on its application within insurance companies, where the authenticity of images is crucial for automated processing in claims management. We provide evidence of ResBuilder's capability to successfully solve the NAS task using our default hyperparameters.
\end{itemize}

The remainder of this work is structured as follows. In \cref{sec:RelatedWork} we discuss the related work and position ourselves in the field of neural architecture search. In \cref{sec:methods} we describe the methodology behind ResBuilder. This is followed by numerical experiments in \cref{sec:results} and finally a conclusion in \cref{sec:discussion_outlook}.

\section{Related Work}
\label{sec:RelatedWork}

AutoML is an active field of research.
While the survery \cite{elsken2019neural} approaches NAS focussing on its different structural aspects, such as comparing different works regarding their search spaces, search strategy and performance estimation on cell structured network architectures where each cell can be arbitrary connected to any of the previous cells and each different cell type stands for a different kind of layer.

The survey \cite{he2021automl} provides an overview of NAS approaches, providing a general and comprehensive view on the field of AutoML. The presented methods are classified into different categories, such as reinforcement learning, evolution-based algorithms, as well as gradient descent, random search and surrogate model-based optimization. Regarding NAS and its performance on image classification tasks, they compare the different methods especially with respect to their accuracy on the CIFAR10 and ImageNet datasets. 

\textbf{NAS and Reinforcement Learning.} The authors of \cite{zoph2016neural} present an approach where they use reinforcement learning in order to predict hyperparameters like the number of filters, filter height/width and stride for every layer up to a chosen maximum of layers using a recurrent neural network as meta model. Another reinforcement learning approach for NAS is introduced in \cite{zoph2018learning} where the authors develop the NASNet search space, in which for a given problem (here CIFAR10 and ImageNet) a certain number of pooling layers are given, between which any number of feature map size-preserving blocks can be inserted. This search space is then iterated using a controller recurrent neural network to design a problem-specific architecture, while our search strategy is simpler and based on a penalization appraoch.

\textbf{Penalization-based NAS.}
Our work is also very related to MorphNet \cite{Gordon17} which optimizes the number of channels of convolutional layers. A group Lasso regularization term penalizes the weights, such that channels with weights below a chosen threshold are removed. The threshold and the penalization are chosen such that the computational cost is below a given budget. In an iterative fashion, this step alternates with a expansion step wherein remaining computational budget is re-distributed proportionally to the different layers of the network. 

Similar to the regularisation terms of BatchNormalisation in MorphNet, gatekeeper variables are introduced in \cite{wang2020pruning} that are multiplied by the output of each channel of each layer. The peculiarity here is that the actual weights are not trained during the process of finding the gatekeeper variables, but are still in the state of their random initialisation, whereby the importance of the layers is calculated before the actual training of the weights starts. To keep the gatekeeper variables low, an additional regularisation term is introduced depending on the cost of the layer and the status of the gatekeeper variables.

Another penalization-based appraoch regarding NAS can be seen in \cite{dong2019network} where the authors present their Transformable Architecture Search (TAS). Instead of the approach of alternating training and pruning of a network architecture, the authors consider the possibility of training a large network with an excessive number of weights and then transferring the knowledge learned to another network for which the depth and width have been determined independently of the first network. The loss function also contains a penalty term that penalises the size of the resulting network.

\textbf{NAS and ResNets.}
ResBuilder works on the search space of ResNet architectures \cite{he2016deep}. In \cite{ahmed2018maskconnect}, the authors also focus on ResNets, in particular the links between the different residual blocks, and define masks such that (in case of ResNet) the input of each residual block can take the outputs of every previous residual block.

The authors of \cite{geifman2019deep} have a similar idea of a modular architecture search space with residual structures as we do, but develop an active-learning approach "incremental neural architecture seatch (iNAS)", which can be combined with any query strategy. They then interpret their search space as a directed acyclic graph in which they start with the smallest possible architecture in order to find a suitable, problem-specific residual architecture.

In contrast to most of these examples, our ResBuilder can be adapted to new data sets or tasks quickly, easily and without adapting further hyperparameters, which simplifies its use for small and medium-sized companies.

\section{Methods}
\label{sec:methods}
ResBuilder operates on the search space of ResNet architectures\cite{he2016deep}. It modulates a given ResNet by optimally dropping layers while randomly inserting layers. At the same time, it is able to shrink and expand the number of filters per layer. For the latter, we utilize MorphNet which relies on a group Lasso regularization term, where the groups refer to weights corresponding to a given channel, as outlined in the previous section. We now introduce the MorphNet algorithm and afterwards wrap our architecture search method around it. 
Our notation is inspired by \cite{Gordon17}.

For a given image of size $H\times W$, let ${F: \mathbb{R}^{H\cdot W} \to \mathbb{R}^{1}}, H,W\in \mathbb{N}$ be a network that assigns class labels to Images:
$$
F = \text{SM} \circ \text{FC} \circ f \circ \text{CL}   
$$

for CL a convolutional, FC a fully connected and SM a softmax layer as well as a backbone ${f: \mathbb{R}^{d_1} \to \mathbb{R}^{d_{2L}}}, $ ${d_1, d_{2L} \in \mathbb{N}}$, $L \in \mathbb{N}$, which consists of $\ell=1,\ldots,L$ ResNet blocks $B_\ell$, i.e.,
$$
f = B_L \circ B_{L-1} \circ \ldots \circ B_1 \,
$$
It should be noted that also pooling layers can be part of the architecture, which are positioned in between of two residual blocks of convolutional layers. However, these are omitted from the notation for the sake of simplicity.

Although our method, in principle, is able to handle residual blocks of other sizes, in this paper we only consider the possibility of inserting a block that contains two convolutional layers as they are used in smaller types of residual networks (e.g. ResNet18, ResNet34\cite{he2016deep}) and are shown as the green block of layers in \cref{fig:archChange_inserted}. Because of this, we consider that each $B_\ell: \mathbb{R}^{d_{2\ell-2}} \to \mathbb{R}^{d_{\ell}} $ is composed of two convolutional layers with weight tensors $\theta_j$, $j \in \{ 2\ell-1, 2\ell \}$. 
For $x_{\ell-1} \in \mathbb{R}^{d_{\ell-1}}$, the operations of these layers are described as
$$
    x_{\ell} = x_{\ell-1} + \sigma( BN(\theta_{2\ell} \cdot \sigma( BN(\theta_{2\ell-1} \cdot x_{\ell-1} ) ) ) )
$$
where $\sigma$ is the ReLU activation $\sigma(t) = \max\{0,t\}$ and $BN$ is the batch normalization process:

$$\displaylines{
BN(z_{i,j,\cdot}) = \left(\frac{z_{i,j,\cdot}-m_{\ell}(z)}{s_{\ell}(z)}\right)\gamma_{\ell}+\beta_{\ell}, z \in \mathbb{R}^{u_{\ell} \times v_{\ell} \times c_{\ell}} \cr
\forall i \in \{1,\ldots, u_{\ell}\},\forall j \in \{1,\ldots,v_{\ell}\} \forall \ell=1,\ldots,2L \cr
}$$

with $\gamma_{\ell},\beta_{\ell}\in\mathbb{R}^{c_l}$ and the mean (standard deviation) $m_{\ell}(z)$ $ (s_{\ell}(z))$ of the tensor $z$ alongside its channels $c_{\ell}$. When $BN$ is applied to the tensor $z$, it is done by applying it to each $z_{i,j,\cdot} \forall i,j$ seperatly.

Any of the $j=1,\ldots,2L$ convolutions maps from $\mathbb{R}^{d_j}$ to $\mathbb{R}^{d_j+1}$, where the dimension is the product $d_j = u_j v_j c_j$ of the spatial dimensions $u_j,v_j$ and the number of channels $c_j$. The application of the $j$th convolution requires 
$$ C_j = 2 s_j u_j v_j c_j c_{j+1} $$
floating point operations, where $s_j = \mathrm{size(\theta_j)} / (c_j c_{j+1}) $ denotes the filter size of the convolution's kernel.

The MorphNet regularization term $\mathcal{G}_M(j)$ for a given layer $j$ is defined as:
\begin{equation}
\begin{split}
    \mathcal{G}_M(\theta, j) = C_j \sum\limits_{p=1}^{c_{j-1}} |\gamma_{j-1, p}| \sum\limits_{k=1}^{c_j} \mathds{1}_{\{\gamma_{j,\cdot} > \tau_M\}}^O + \\
    C_j \sum\limits_{p=1}^{c_j}\mathds{1}_{\{\gamma_{j-1} > \tau_M\}}^I \sum\limits_{k=1}^{c_{j-1}}|\gamma_{j,p}|
\end{split}
\end{equation}
with $\mathds{1}^I$ ($\mathds{1}^O$) the indicator function for a given statement on the input (output) channels.
For further information on how the MorphNet regularization term is calculated have a look in \cite{Gordon17}.
The overall MorphNet regularization term $\mathcal{G}_M$ is then defined by:
\begin{equation}
    \mathcal{G}_M = \sum\limits_{\ell=1}^{2L}\mathcal{G}_M(\theta, \ell)
\end{equation}

The Morphnet algorithm (presented in \cite{Gordon17}) is given in \cref{alg:MorphNet}, where $c_{\ell}'$ denotes the channel width of Layer $\ell$, $\mathcal{F}$ is in our case the number of floating point operations (FLOPs) the net currently uses and $\zeta$ the maximum capacity of FLOPs the net should have. Furthermore, we denote $c_{1:2L} = (c_1,\ldots,c_{2L}) $.

\begin{algorithm}
\caption{The MorphNet algorithm}\label{alg:MorphNet}
    \begin{algorithmic}[1]
        \State Train the network to find $$\theta^{*} = \argmin\theta \{\mathcal{L}(\theta) + \lambda_M \mathcal{G}_M(\theta)\} $$  for suitable $\lambda_M$.
        \State Find the new widths $c_{1:2L}'$ induced by $\theta^*$.
        \State Find the largest $\omega$, such that $\mathcal{F}(\omega \cdot c_{1:2L}') \leq \zeta$.
        \State Repeat from Step 1 for as many times as desired, setting $c_{1:2L}^0 = \omega \cdot c_{1:2L}'$.
    \end{algorithmic}
\end{algorithm}

Our LayerLasso method also uses a $L1$-regularization term $\mathcal{G}_\Lambda$ which is based on regularization of the weights $\theta$:

\begin{equation}
    \mathcal{G}_\Lambda = \sum\limits_{\ell = 1}^{2L} \|\theta_{\ell}\|_1 .
\end{equation}

It should be noted that this regularisation term is only applied to layers in the residual blocks and does neither penalise the weights of the initial layer nor the fully connected layers.

Besides the regularization terms, our loss function also consists of a cross entropy loss $\mathcal{L}_{\text{CE}}$:

\begin{equation}
    \mathcal{L}_{\text{CE}} = -\sum\limits_i^q y_i\log F_i(x).
\end{equation}

In summary, our loss function consists of three different parts:
\begin{itemize}
    \item The weight-optimizing part $\mathcal{L}_{\text{CE}}$ including a default L2-regularization term 
    \item The MorphNet-L1-regularization $\mathcal{G}_{M}$ multiplied by the MorphNet regularization strength $\lambda_{M}$
    \item The LayerLasso-L1-regularization $\mathcal{G}_{\Lambda}$ multiplied by the MorphNet regularization strength $\lambda_{\Lambda}$
\end{itemize}
The complete loss term $\mathcal{L}$ can then be defined as:

\begin{equation}
    \mathcal{L} = \mathcal{L}_{\text{CE}}+\lambda_{\Lambda} \mathcal{G}_{\Lambda} + \lambda_M \mathcal{G}_M
\end{equation}

\subsection{Inserting ResNet blocks}
\begin{figure*}
\begin{subfigure}{0.26\linewidth}
  \centering
  \includegraphics[height=0.15\textheight]{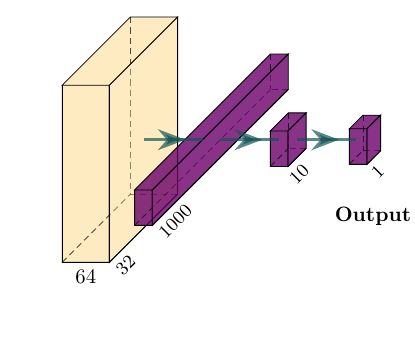}
  \caption{An empty start net with only one convolutional layer.}
  \label{fig:archChange_emptynet}
\end{subfigure}
\hspace{0.049\linewidth}
\begin{subfigure}{0.26\linewidth}
  \centering
  \includegraphics[height=0.15\textheight]{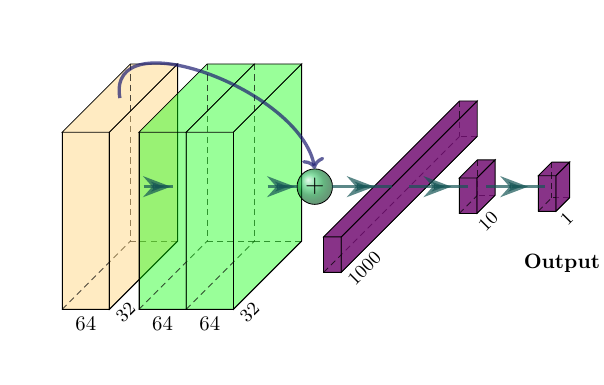}
  \caption{The net from \Cref{fig:archChange_emptynet} after inserting a block of two convolutional layers (green).}
  \label{fig:archChange_inserted}
\end{subfigure}
\hspace{0.049\linewidth}
\begin{subfigure}{0.26\linewidth}
  \centering
  \includegraphics[height=0.15\textheight]{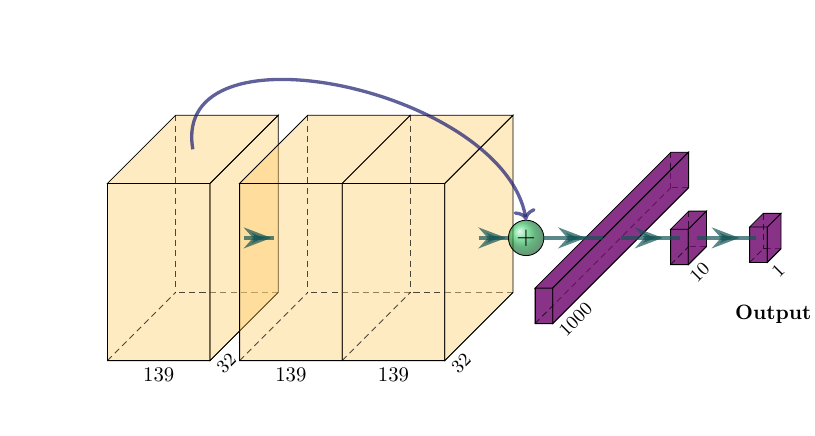}
  \caption{The net from Fig. \Cref{fig:archChange_inserted} after one MorphNet iteration.}
  \label{fig:archChange_morphed}
\end{subfigure}%
\caption{Example of structural changes the \textit{ResBuilder} method uses.}
\label{fig:archChange_full}
\end{figure*}

\begin{figure}
    \centering
    \includegraphics[width=\linewidth]{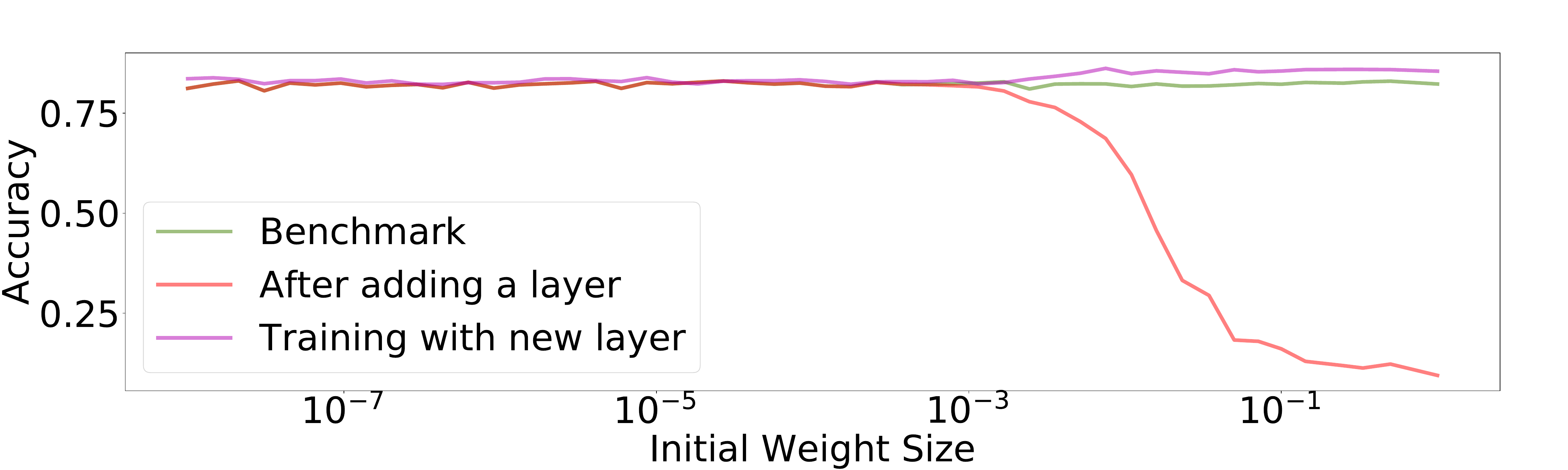}
    \caption{Effect of weight initialization of an inserted layer block to a network on the CIFAR10 dataset. The green line indicates the benchmark of the startnet trained normally without a new block of layers inserted. Red indicates the accuracy after adding one of these blocks to our startnet but without further training while violet shows the accuracy after further training with the new layerblock inserted.}
    \label{fig:weight_init}
\end{figure}

Due to the construction of ResNet blocks, in particular their identity given by the addition of $x^{(B_i)}$, it is natural to initialize $\theta^{(B_i,j)}$, $j=1,2$, close to zero (but randomly) in order to enable a seamless continuation of training from the current state of the network. Initialization close to zero implies $ x^{(B_i+1)} \approx x^{(B_i)} $. Randomness is required to avoid symmetries in the weights that occur in case of zero initialization.

In order to obtain the most effective trade-off between an initialisation close to 0 and the avoidance of undesired symmetries in the network's weights, we show in \cref{fig:weight_init} how much the accuracy of the network suffers from the insertion of residual blocks with different initial weights, but also how strong the constraints of unbroken symmetries would be when continuing to train.

Therefore, we insert a layer block $B_k$ behind a given block $B_i$ into our network $f$ according to our insertion strategy (see \cref{subsec:strat_ins_del}) like it can be seen in \cref{fig:archChange_full}. We choose its initial weights such that the full potential of the new block can be exploited but the damage to the current knowledge of the network remains minimal. In our example, we therefore initialize the weights with an average initial weight of the order of $\theta_{init}=10^{-2}$. From this, we get the new network architecture: $$f'=B_n \circ \ldots \circ B_{i+1} \circ B_k \circ B_{i} \circ \ldots \circ B_1$$

\subsection{LayerLasso}
\label{subsec:LayerLasso}

In order to also have the possibility to delete residual blocks of layers from positions where the net does not use its capacity efficiently, we introduce the \textbf{LayerLasso}:
After a certain number of epochs of training every block $B_i$ that includes at least one layer whose sum of weights $\sum_{\theta_m \in \theta^{B_i,j}} \|\theta_m\|_1, j=1,2$ lays under the set threshold for layer deleting $\tau_{\Lambda}$ will be erased from the net: 
\begin{equation}
\begin{split}
f' = B_n \circ \ldots \circ B_{i+1} \circ B_{i-1} \circ \ldots \circ B_1 \\
\forall B_i \exists j: \sum_{\theta_m \in \theta^{B_i,j}} \|\theta_m\|_1 < \tau_{\Lambda} .    
\end{split}
\end{equation}

Because of the residual architecture, a layer with with small weights resembles the identity, so we continue training after the removal of blocks from the resulting architecture with the weights for the remaining blocks equal to the values before the removal step. 

\subsection{Strategy of inserting and removing}
\label{subsec:strat_ins_del}
In order to optimize the resulting network architecture we used a "random in - greedy out" optimization strategy: We insert residual blocks of convolutional layers at random positions in the net with the only constraints that the new block can not be inserted before the first convolutional layer or after the flattening of the feature maps. In order to evenly spread the blocks across the entire architecture we choose the pooling stage randomly and then a random block $B_i$ from this stage after which a new residual block is inserted (this might also be directly behind the pooling layer).

\subsection{Structure of the method}
\label{subsec:structure_pipeline}
\begin{figure}
    \centering
    \includegraphics[width=\linewidth]{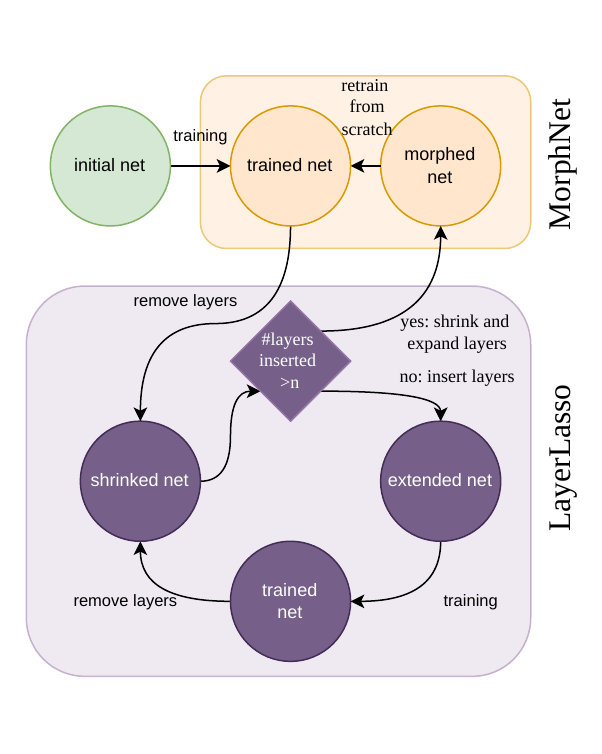}
    \caption{Overview of the method}
    \label{fig:trainingspipeline}
\end{figure}

Our NAS training pipeline thus is defined as depicted in \Cref{fig:trainingspipeline}. We do $n_{\Lambda}$ insertion steps before we start the MorphNet subroutine. Besides the training with all regularization terms active

$\theta_{reg}:= \argmin {} \{\mathcal{L}_{\text{CE}}(\theta) + \lambda_{M}\mathcal{G}_{M} + \lambda_{\Lambda}\mathcal{G}_{\Lambda}\} $

our training pipeline determines which elements $\mathcal{A}'\subset\mathcal{A}$ of the architecture search space $\mathcal{A}$ are considered, while we also train the actual network architecture without regularization 
$\theta_{noReg}:=$ $\argmin{} \{\mathcal{L}_{\text{CE}}\}$ 
in order to achieve the best accuracy for the given net.

\section{Numerical experiments}
\label{sec:experimental_setup}
In this section we give an overview of our experimental setup and provoide hyperparameters we use. In this section we also present our numerical results on six academic data sets and for one real world industrial application.

\subsection{Experimental setup}
\label{subsec:experimental_setup}

\begin{figure}
    \centering
    \includegraphics[width=0.8\linewidth, trim=20 20 0 0, clip]{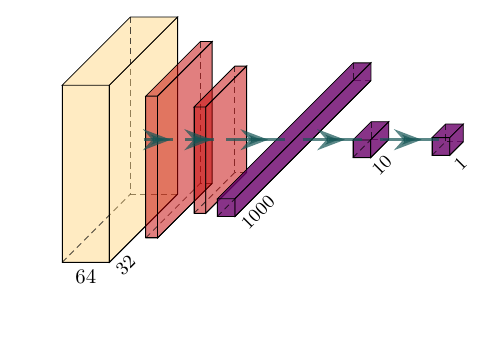}
    \caption{This is the minimal initial startnet architecture with one convolutional layer (orange), two pooling layers (red) and one dense layer (violet) before the softmax and output layer.}
    \label{fig:emptyStarnetArchitecture}
\end{figure}

\paragraph{Initial architectures}
\label{subsec_experimental_setup_used_architectures}
In our experiments, we apply our ResBuilder method to the different data sets. In one set of experiments, a ResNet18 is used as the initial network (RB-R18), and in the second set, a minimal network (RB-0Net), as shown in \cref{fig:emptyStarnetArchitecture}. 

\paragraph{Training types}
The experiments we conduct contain three training types of each neural network architecture that is considered: 
\begin{itemize}
    \item Training Variant \textit{With Reg}: With all possible regularisation terms. 
    \item Training Variant \textit{No Reg RI}: Without additional regularisation terms, starting from scratch, i.e. a random initialisation of the weights.
    \item Training Variant \textit{No Reg WI}: Without additional regularisation terms, starting from the checkpoint induced by training \textit{With Reg}.
\end{itemize}
The expression "without additional regularisation terms" used here means that the training takes place without additional regularisation by MorphNet or the LayerLasso, i.e. $\lambda_M = \lambda_{\Lambda} = 0$. An additional $L2$-regularisation term $\lambda_0$, which is intended to reduce overfitting, is nevertheless applied to the training.

\paragraph{Adjustable Hyperparameters}
Unless further specified, we use the default values given here for the hyperparameters:
\begin{itemize}
    \item $n_{\Lambda}=4$ number of insertion steps before MorphNet step.
    \item $n_M = 7$ number of MorphNet steps.
    \item $\lambda_M = 10^{-7}$ the MorphNet regularization strength
    \item $\zeta = 100{,}000{,}000$ the aimed for FLOP costs, except for Animals10, for which ten times the amount of arithmetic operations was allowed ($\zeta = 1{,}000{,}000{,}000$), as the images in this dataset also have significantly higher resolution.
    \item $\lambda_{\Lambda} = 10^{-8}$ the LayerLasso regularizaion strength
    \item $\tau_{\Lambda} = 10^{-3}$ the threshold of our LayerLasso method which sets when a layer (block) is deleted (see \cref{subsec:LayerLasso})
    \item $\lambda_{0} = 10^{-5}$ the additional $L2$-regularization strength 
    
    \item \textit{Adam} is used as optimizer.
    \item Data augmentation is active with 10\% shifts in horizontal and vertical position as well as a possible horizontal flip.
\end{itemize}

\paragraph{Academic datasets}
\label{subsec:datasets}
In this work we evaluate our methods on the datasets listed in \cref{tab:datasets_overview}: 

\begin{table}[h]
    \centering
    \begin{center}
        \begin{tabular}{|c|c|c|c|c|}
            \hline
            Name & Content & Size (px) & Classes\\
            \hline
            Animals10 \cite{animals10dataset} & Animals & 300 & 10\\ 
            \hline
            CIFAR10 \cite{CIFAR10} & Misc. & 32 & 10\\
            \hline
            CIFAR100 \cite{CIFAR10} & Misc. & 32 & 100\\
            \hline
            MNIST \cite{lecun1998MNIST} & Digits & 28 & 10\\ 
            \hline
            FashionMNIST \cite{FashionMNIST} & Clothing & 28 & 10\\
            \hline
            EMNIST \cite{cohen2017emnist} & Letters & 28 & 62\\ 
            \hline
        \end{tabular}
    \end{center}
\caption{Overview of datasets we use in this work.}
\label{tab:datasets_overview}
\end{table}

\paragraph{\textbf{Preprocessing the Data}}
\label{subsec:MaM_Preprocessing}
For the Animals10 and CIFAR10 datasets, we apply a normalisation process to the data, which calculates for each pixel the difference between the current pixel value and the mean of all pixel values of all images in the current colour channel, and then divides it by the standard deviation.

MNIST and FashionMNIST pixel values are simply scaled from 0 to 1 and not pre-processed.

We do also use data augmentation for our trainings, where we allow a shift of 10\% in both dimensions and also horizontal flips.

\paragraph{\textbf{Used Ressources}}
\label{subsec:used_software}
We used different packages like Tensorflow-GPU (version 1.14.0) \cite{tensorflow2015-whitepaper}, Keras (version 2.2.4.) \cite{chollet2015keras} or MorphNet (0.2.1) \cite{Gordon17}. 
The visualization of neural nets like \cref{fig:emptyStarnetArchitecture} is based on the repository of \cite{haris_iqbal_2018_2526396}. 
For a full list of used packages see the requirements.txt in our git repository. 
\label{subsec:hardware}
For the calculations, we used a Dell Precision 7920 workstation with a Dual Intel Xeon Gold 6248R 3.0GHz and three Nvidia Quadro P 6000 graphic units with 24GB VRAM each.

\subsection{Numerical Results}
\label{sec:results}

\paragraph{\textbf{Results on academic datasets}}

\begin{table}
\centering
    \begin{tabular}{ |c|c|c|c|c|  }
    \hline
    Dataset & Res18 & MorphNet & RB-R18 & RB-0Net\\
    \hline
    Animals10 & 92.10\% & $92.02\%^*$ & \textbf{92.73\%}$^*$ & $88.72\%^*$\\
    CIFAR10 & 85.50\% & 88.17\% & 88.32\% & \textbf{89.92\%} \\
    CIFAR100 & 53.80\% & 59.78\% & 57.69\% & \textbf{62.36\%} \\
    MNIST & 99.14\% & 99.11\% & 99.17\% & \textbf{99.34\%}\\
    FashionMNIST & 92.81\% & 92.97\% & \textbf{93.55\%} & 93.71\% \\
    EMNIST & 86.48\% & 86.70\% & 86.86\% & \textbf{86.95\%} \\
    \hline
    \end{tabular}
\caption{Overview of achieved accuracies}
\label{tab:results_acc}
\end{table}

In one of our experimental setups, we ran the ResBuilder method with the same hyper-parameters (described in \cref{sec:experimental_setup}) for all our datasets under consideration and summarise these results in \cref{tab:results_acc}.
The column with the results of the run "RB-0Net" gives the accuracies we achieve for the test series starting with a minimal network architecture as shown in \cref{fig:emptyStarnetArchitecture}, without regularisation (\textit{No Reg RI} and \textit{No Reg WI}), as this would be the later use case.
The results in "RB-R18" were generated analogously to the results of "RB-0Net", with the exception that a ResNet18 \cite{he2016deep} was used as the initial architecture.
We benchmark our method against the accuracy we achieve by training a ResNet18 on the specific dataset. The column "Res18" therefore gives the accuracy when it is trained with the variant "No Reg RI" which would be the typcial way to go in order to train this architecture from scratch.
The "MorphNet"-column shows an ablation study where we omit our additional LayerLasso routine from the architecture search process wherefore we use a ResNet18 and perform the single MorphNet routine for the given number of $n_M$. The training that is used for that, follows the variant "No Reg RI", as our Res18 benchmark also does.

As can be seen in \cref{tab:results_acc}, our ResBuilder method performs better on all datasets than both the standard ResNet18 architecture and the MorphNet approach without our depth-first search.

\begin{figure}[h]
    \centering
    \includegraphics[width=\linewidth]{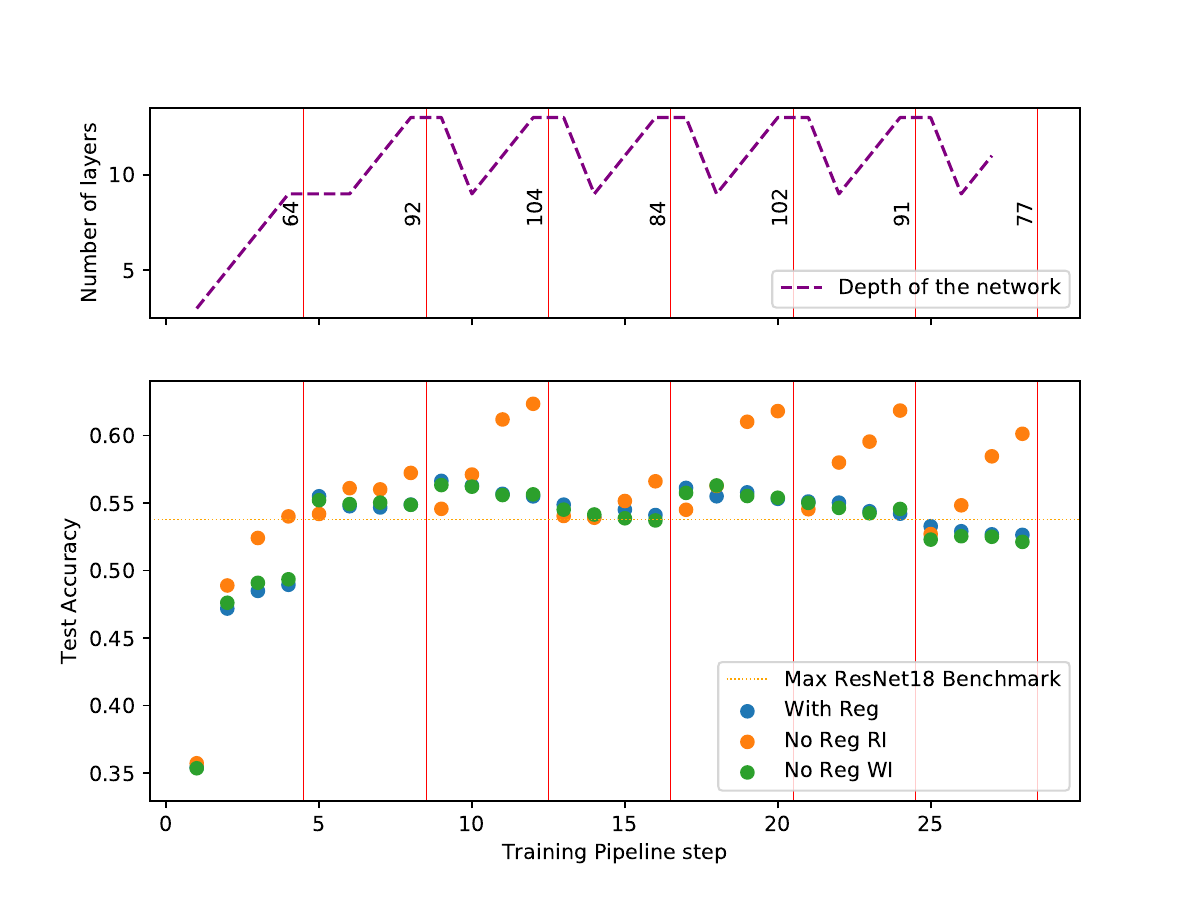}
    \caption{Accuracies of different architectures on CIFAR100 started with an empty net (see \cref{fig:emptyStarnetArchitecture}) as initial net.}
    \label{fig:Acc_CIF100_fromEN}
\end{figure}

\begin{figure*}[h]
    \centering
    \includegraphics[width=\linewidth]{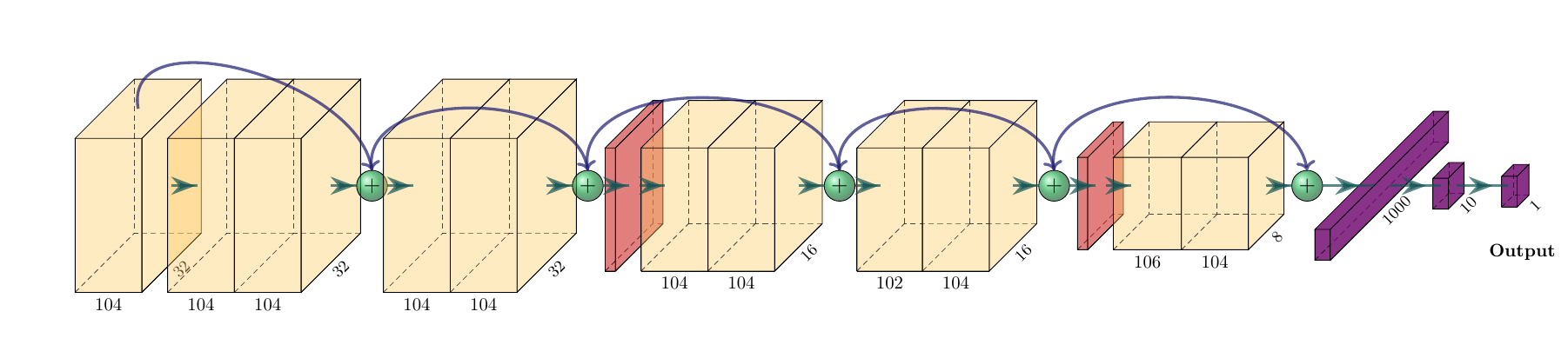}
    \caption{Best architecture (pipeline step 12) from \cref{fig:Acc_CIF100_fromEN}.}
    \label{fig:arch_CIF100_fromEN}
\end{figure*}

\begin{figure}[h]
    \centering
    \includegraphics[width=\linewidth]{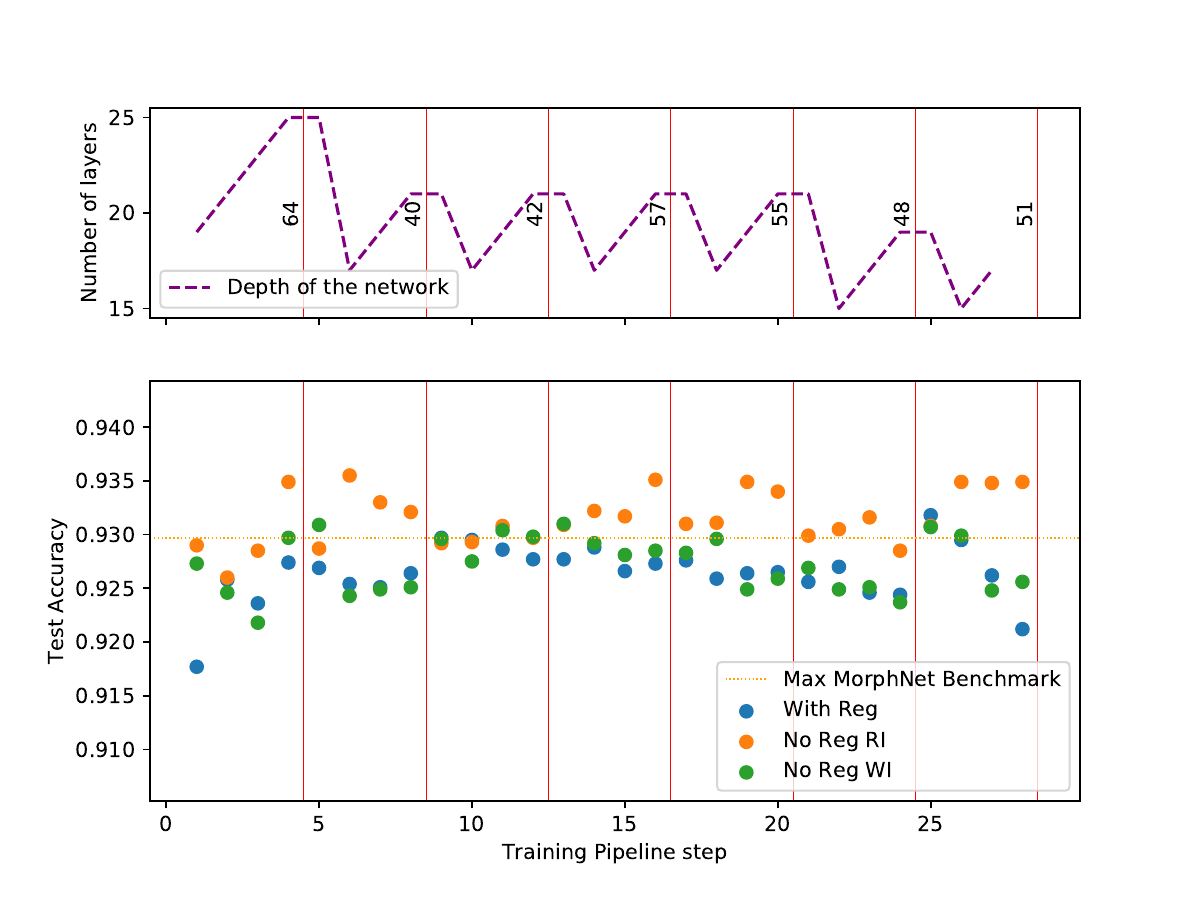}
    \caption{Accuracies of different architectures on FashionMNIST started with a ResNet18 architecture as initial net.}
    \label{fig:Acc_FashMNIST_fromRes18}
\end{figure}

\begin{figure*}[h]
    \centering
    \includegraphics[width=\linewidth]{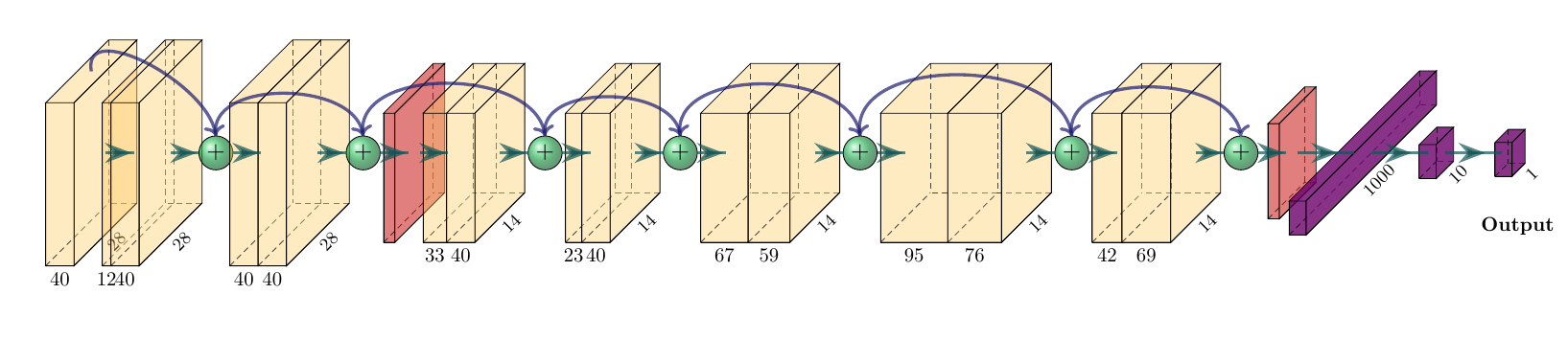}
    \caption{Best architecture (pipeline step 6) from \cref{fig:Acc_FashMNIST_fromRes18}.}
    \label{fig:Arch_FashMNIST_fromRes18}
\end{figure*}

In \cref{fig:Acc_CIF100_fromEN} we can see in the lower panel how the accuracy for the three different training types over the iterations of architectures in the search space (shown on the x-axis). Here we start searching from the minimal initial architecture (\cref{fig:emptyStarnetArchitecture}) on the CIFAR100 dataset. Meanwhile, the upper part of the figure shows how the depth of the network architecture while our ResBuilder method progresses, which is represented by the purple dashed line. Vertical red lines within the figure indicate MorphNet channel width optimization steps and the numbers next to the line in the upper part of the  display the maximum channel size in the first pooling block in order to get a feeling for the width of the layer.
Since this figure shows an experiment that started with a minimal initial architecture, the orange line shows the benchmark that was achieved by training a ResNet18. It should be noted that even at a fairly early stage of the ResBuilder, accuracies above this benchmark could be achieved. 
In this experiment, the best accuracy was achieved with the twelfth architecture, which is therefore shown in \cref{fig:arch_CIF100_fromEN}.
Furthermore, one can see that both the accuracy achieved and the depth and width of the network have converged very quickly in this example.

Another example is given in \cref{fig:Acc_FashMNIST_fromRes18}, which is similar to \cref{fig:Acc_CIF100_fromEN} except for the orange benchmark line, which is here the accuracy of the best MorphNet run as this run already started at the ResNet18 as initial architecture and so this benchmark can be seen from the first "column" ($x=1$) of the figure. One other difference is the used dataset which is FashionMNIST in this case. Here it is nice to see that the size of the network in terms of both depth and width can be steadily reduced without losing accuracy, thus saving computing resources compared to ResNet18. The best performing architecture is also shown for this example in \cref{fig:Arch_FashMNIST_fromRes18}. This may be in part due to the fact that FashionMNIST, like many of the MNIST datasets, can also be predicted well by small or cost-effective networks.


In the appendix, \cref{fig:archis_CIF_Res18_1}, \cref{fig:archis_CIF_Res18_2} and \cref{fig:archis_CIF_Res18_3} exemplify a complete history of the architectures during a run of the ResBuilder method, starting from ResNet18 on the CIFAR10 dataset. There you can follow the individual insertion/elimination and MorphNet steps, whereby newly inserted convolutional layers are always marked green and the widths of the individual convolutional layers can be seen in their number underneath each yellow block.

\paragraph{\textbf{Removal positions}}
\label{subsec:removal_pos}

\begin{figure}[h]
    \centering
    \includegraphics[width=\linewidth]{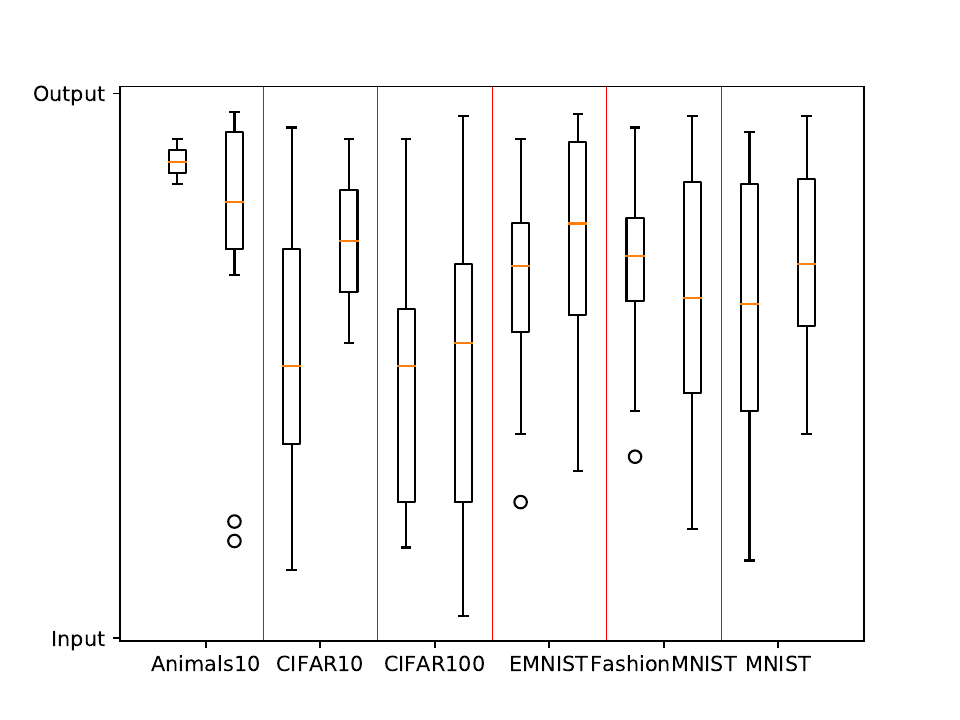}
    \caption{Position of removed layerblocks relativ to the network size. For each dataset the left plot shows the removed positions proceeding from the minimal initial architecture and the right one proceeding from the ResNet18 architecture.}
    \label{fig:removed_positions}
\end{figure}

In \cref{fig:removed_positions}, we show at which positions of the network our method eliminates blocks of layers. The positions between input and output are always to be seen relative to the current network size. 
In particular, it can be observed that for CIFAR100, which has a relatively large number of classes (100), more layers tend to be thrown out of the front part of the mesh, which could result in a focus on the classification into the individual classes instead of optimizing encoding. The rather higher resolution of the Animals10 dataset, on the other hand, tends to throw out layer blocks at the back end of the architecture, which speaks for the importance of the front blocks for extracting information from the high resolution images.

\paragraph{\textbf{Regularization parameter study}}
\label{subsec:regstr}

\begin{figure}[h]
    \centering
    \includegraphics[width=\linewidth]{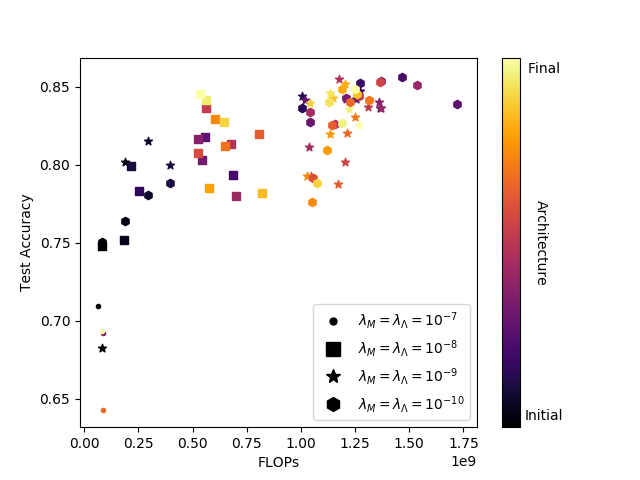}
    \caption{Different regularization strengths for training on CIFAR10 data.}
    \label{fig:accvsflops_regulruns}
\end{figure}

\Cref{fig:accvsflops_regulruns} shows the accuracies of training different architectures with various regularization strengths. To ensure better visualisation, only the experiments with $\lambda_M = \lambda_{\Lambda}$ are considered in the figure. The shown accuracies all refer to the training variant \textit{With Reg} without additional L2-regularization ($\lambda_0 = 0$). One can see that there is an accuracy trade-off between low FLOPs induced by a high regularization term and high FLOPs with a low regularization strength. For $\lambda_M = \lambda_{\Lambda} = 10^{-7}$ it can be also mentioned that the architecture broke down due to too high penalization applied.

\paragraph{\textbf{Example of industrial application}}

In addition, the ResBuilder methodology has been challenged not only on standard benchmark datasets, but also on an independent set of data to assess its applicability in real-world scenarios. Specifically, the approach was applied to an insurance use case offered by ControlExpert (CE)\cite{ControlExpert}, a company providing end-to-end motor claims management solutions. CE uses images, in particular of damaged vehicles, to process claims. For this, it must be ensured that these images are authentic, for instance by detecting manipulated images. This process can be described as a binary classification task. Ordinarily, a variety of architectures are scrutinized in these tasks to determine the optimal fit. These architectures are quite sophisticated, but are generally not developed specifically for a single classification task. Additionally, conducting the experiments to find an optimized architecture is time consuming. Hence, in an industry research process this might lead to competitive disadvantage. This motivated the application of the ResBuilder approach and its comparison with competitor networks like EfficientNet-b0\cite{tan2019efficientnet}, EfficientNet-b4\cite{tan2019efficientnet}, and ResNet18\cite{he2016deep}. The dataset utilized for this experiment was balanced, comprising 50\% manipulated images and 50\% non-manipulated images of damaged vehicles. The image manipulations were performed manually using various forgery methods such as splicing and copy-move. ResBuilder demonstrated a superior performance in terms of accuracy, achieving an enhancement of 1.2\%-points\footnote{Absolute values cannot be disclosed due to ControlExpert's intellectual property rights.} compared to the most effective competitor network, EfficientNet-b0. Moreover, it attained this while maintaining a reduction of 97.86\% in parameter quantity. The findings suggest that ResBuilder is capable of generating efficient architectures for a real-world scenario, exhibiting not only an improved accuracy for the use case but also a significant memory efficiency. Another benefit of the ResBuilder approach is its capability to automate the architecture search process, thus reducing the manual efforts required in creating an efficient model, a factor of substantial importance in an industrial context in order to lower the time-to-market for the development of such AI models.

\section{Discussion}
\label{sec:discussion_outlook}

In this paper, we introduced the ResBuilder method, which provides a NAS algorithm that can generate neural networks from scratch or from existing architectures using suitable regularisation techniques. On many datasets (mainly academic, but also industrial), results close to state-of-the-art (without pretraining) are achieved. To this end, ablation studies related to the omission of our LayerLasso method were conducted and a parameter study on the regularisation strength was performed.

\printbibliography

@misc{tensorflow2015-whitepaper,
title={ {TensorFlow}: Large-Scale Machine Learning on Heterogeneous Systems},
url={https://www.tensorflow.org/},
note={Software available from tensorflow.org},
author={
    Mart\'{\i}n~Abadi and
    Ashish~Agarwal and
    Paul~Barham and
    Eugene~Brevdo and
    Zhifeng~Chen and
    Craig~Citro and
    Greg~S.~Corrado and
    Andy~Davis and
    Jeffrey~Dean and
    Matthieu~Devin and
    Sanjay~Ghemawat and
    Ian~Goodfellow and
    Andrew~Harp and
    Geoffrey~Irving and
    Michael~Isard and
    Yangqing Jia and
    Rafal~Jozefowicz and
    Lukasz~Kaiser and
    Manjunath~Kudlur and
    Josh~Levenberg and
    Dandelion~Man\'{e} and
    Rajat~Monga and
    Sherry~Moore and
    Derek~Murray and
    Chris~Olah and
    Mike~Schuster and
    Jonathon~Shlens and
    Benoit~Steiner and
    Ilya~Sutskever and
    Kunal~Talwar and
    Paul~Tucker and
    Vincent~Vanhoucke and
    Vijay~Vasudevan and
    Fernanda~Vi\'{e}gas and
    Oriol~Vinyals and
    Pete~Warden and
    Martin~Wattenberg and
    Martin~Wicke and
    Yuan~Yu and
    Xiaoqiang~Zheng},
  year={2015},
}

@misc{chollet2015keras,
  title={Keras},
  author={Chollet, Fran\c{c}ois and others},
  year={2015},
  howpublished={\url{https://keras.io}},
}

@article{FashionMNIST,
  author    = {Han Xiao and
               Kashif Rasul and
               Roland Vollgraf},
  title     = {Fashion-MNIST: a Novel Image Dataset for Benchmarking Machine Learning
               Algorithms},
  journal   = {CoRR},
  volume    = {abs/1708.07747},
  year      = {2017},
  url       = {http://arxiv.org/abs/1708.07747},
  eprinttype = {arXiv},
  eprint    = {1708.07747},
  bibsource = {dblp computer science bibliography, https://dblp.org}
}

@article{CIFAR10,
  title={Learning multiple layers of features from tiny images},
  author={Krizhevsky, Alex and Hinton, Geoffrey and others},
  year={2009},
  publisher={Citeseer}
}

@article{lecun1998MNIST,
  title={Gradient-based learning applied to document recognition},
  author={LeCun, Yann and Bottou, L{\'e}on and Bengio, Yoshua and Haffner, Patrick},
  journal={Proceedings of the IEEE},
  volume={86},
  number={11},
  pages={2278--2324},
  year={1998},
  publisher={Ieee}
}

@article{Gordon17,
  author      = {Ariel Gordon and Elad Eban and Ofir Nachum and Bo Chen and Hao Wu and Tien-Ju Yang and Edward Choi},
  title       = {MorphNet: Fast \& Simple Resource-Constrained Structure Learning of Deep Networks},
year={2017},
  date        = {2017-11-18},
  eprint      = {http://arxiv.org/abs/1711.06798v3},
  eprintclass = {cs.LG},
  eprinttype  = {arXiv}
}

@inproceedings{he2016deep,
  title={Deep residual learning for image recognition},
  author={He, Kaiming and Zhang, Xiangyu and Ren, Shaoqing and Sun, Jian},
  booktitle={Proceedings of the IEEE conference on computer vision and pattern recognition},
  pages={770--778},
  year={2016}
}

@article{he2021automl,
  title={AutoML: A survey of the state-of-the-art},
  author={He, Xin and Zhao, Kaiyong and Chu, Xiaowen},
  journal={Knowledge-Based Systems},
  volume={212},
  pages={106622},
  year={2021},
  publisher={Elsevier}
}

@article{russakovsky2015imagenet,
  title={Imagenet large scale visual recognition challenge},
  author={Russakovsky, Olga and Deng, Jia and Su, Hao and Krause, Jonathan and Satheesh, Sanjeev and Ma, Sean and Huang, Zhiheng and Karpathy, Andrej and Khosla, Aditya and Bernstein, Michael and others},
  journal={International journal of computer vision},
  volume={115},
  pages={211--252},
  year={2015},
  publisher={Springer}
}

@article{nakano2021webgpt,
  title={Webgpt: Browser-assisted question-answering with human feedback},
  author={Nakano, Reiichiro and Hilton, Jacob and Balaji, Suchir and Wu, Jeff and Ouyang, Long and Kim, Christina and Hesse, Christopher and Jain, Shantanu and Kosaraju, Vineet and Saunders, William and others},
  journal={arXiv preprint arXiv:2112.09332},
  year={2021}
}

@misc{animals10dataset,
  title = {Animals10 dataset},
  howpublished = {\url{https://www.kaggle.com/datasets/alessiocorrado99/animals10}},
  note = {Accessed: 2022-05-30}
}

@article{elsken2019neural,
  title={Neural architecture search: A survey},
  author={Elsken, Thomas and Metzen, Jan Hendrik and Hutter, Frank},
  journal={The Journal of Machine Learning Research},
  volume={20},
  number={1},
  pages={1997--2017},
  year={2019},
  publisher={JMLR. org}
}

@article{reiners2022efficient,
  title={Efficient and sparse neural networks by pruning weights in a multiobjective learning approach},
  author={Reiners, Malena and Klamroth, Kathrin and Heldmann, Fabian and Stiglmayr, Michael},
  journal={Computers \& Operations Research},
  volume={141},
  pages={105676},
  year={2022},
  publisher={Elsevier}
}

@inproceedings{he2017channel,
  title={Channel pruning for accelerating very deep neural networks},
  author={He, Yihui and Zhang, Xiangyu and Sun, Jian},
  booktitle={Proceedings of the IEEE international conference on computer vision},
  pages={1389--1397},
  year={2017}
}

@article{srinivas2015data,
  title={Data-free parameter pruning for deep neural networks},
  author={Srinivas, Suraj and Babu, R Venkatesh},
  journal={arXiv preprint arXiv:1507.06149},
  year={2015}
}

@article{liu2018rethinking,
  title={Rethinking the value of network pruning},
  author={Liu, Zhuang and Sun, Mingjie and Zhou, Tinghui and Huang, Gao and Darrell, Trevor},
  journal={arXiv preprint arXiv:1810.05270},
  year={2018}
}

@article{choudhary2020comprehensive,
  title={A comprehensive survey on model compression and acceleration},
  author={Choudhary, Tejalal and Mishra, Vipul and Goswami, Anurag and Sarangapani, Jagannathan},
  journal={Artificial Intelligence Review},
  volume={53},
  pages={5113--5155},
  year={2020},
  publisher={Springer}
}

@inproceedings{ahmed2018maskconnect,
  title={Maskconnect: Connectivity learning by gradient descent},
  author={Ahmed, Karim and Torresani, Lorenzo},
  booktitle={Proceedings of the European Conference on Computer Vision (ECCV)},
  pages={349--365},
  year={2018}
}

@inproceedings{cohen2017emnist,
  title={EMNIST: Extending MNIST to handwritten letters},
  author={Cohen, Gregory and Afshar, Saeed and Tapson, Jonathan and Van Schaik, Andre},
  booktitle={2017 international joint conference on neural networks (IJCNN)},
  pages={2921--2926},
  year={2017},
  organization={IEEE}
}

@article{zoph2016neural,
  title={Neural architecture search with reinforcement learning},
  author={Zoph, Barret and Le, Quoc V},
  journal={arXiv preprint arXiv:1611.01578},
  year={2016}
}

@inproceedings{zoph2018learning,
  title={Learning transferable architectures for scalable image recognition},
  author={Zoph, Barret and Vasudevan, Vijay and Shlens, Jonathon and Le, Quoc V},
  booktitle={Proceedings of the IEEE conference on computer vision and pattern recognition},
  pages={8697--8710},
  year={2018}
}

@book{goodfellow2016deep,
  title={Deep learning},
  author={Goodfellow, Ian and Bengio, Yoshua and Courville, Aaron},
  year={2016},
  publisher={MIT press}
}

@misc{haris_iqbal_2018_2526396,
  author       = {Haris Iqbal},
  title        = {HarisIqbal88/PlotNeuralNet v1.0.0},
  month        = dec,
  year         = 2018,
  doi          = {10.5281/zenodo.2526396},
  url          = {https://doi.org/10.5281/zenodo.2526396}
}

@article{dong2019network,
  title={Network pruning via transformable architecture search},
  author={Dong, Xuanyi and Yang, Yi},
  journal={Advances in Neural Information Processing Systems},
  volume={32},
  year={2019}
}

@inproceedings{wang2020pruning,
  title={Pruning from scratch},
  author={Wang, Yulong and Zhang, Xiaolu and Xie, Lingxi and Zhou, Jun and Su, Hang and Zhang, Bo and Hu, Xiaolin},
  booktitle={Proceedings of the AAAI Conference on Artificial Intelligence},
  volume={34},
  number={07},
  pages={12273--12280},
  year={2020}
}

@article{geifman2019deep,
  title={Deep active learning with a neural architecture search},
  author={Geifman, Yonatan and El-Yaniv, Ran},
  journal={Advances in Neural Information Processing Systems},
  volume={32},
  year={2019}
}

@article{hsu2018monas,
  title={Monas: Multi-objective neural architecture search using reinforcement learning},
  author={Hsu, Chi-Hung and Chang, Shu-Huan and Liang, Jhao-Hong and Chou, Hsin-Ping and Liu, Chun-Hao and Chang, Shih-Chieh and Pan, Jia-Yu and Chen, Yu-Ting and Wei, Wei and Juan, Da-Cheng},
  journal={arXiv preprint arXiv:1806.10332},
  year={2018}
}

@article{yao2018taking,
  title={Taking human out of learning applications: A survey on automated machine learning},
  author={Yao, Quanming and Wang, Mengshuo and Chen, Yuqiang and Dai, Wenyuan and Li, Yu-Feng and Tu, Wei-Wei and Yang, Qiang and Yu, Yang},
  journal={arXiv preprint arXiv:1810.13306},
  year={2018}
}

@misc{ControlExpert,
  title = {ControlExpert},
  howpublished = {\url{https://www.controlexpert.com/de-de/}},
  note = {Accessed: 2023-07-05}
}

@inproceedings{tan2019efficientnet,
  title={Efficientnet: Rethinking model scaling for convolutional neural networks},
  author={Tan, Mingxing and Le, Quoc},
  booktitle={International conference on machine learning},
  pages={6105--6114},
  year={2019},
  organization={PMLR}
}
\appendix

\begin{figure*}
    \begin{subfigure}{\textwidth}
        \centering
        \includegraphics[width=\linewidth]{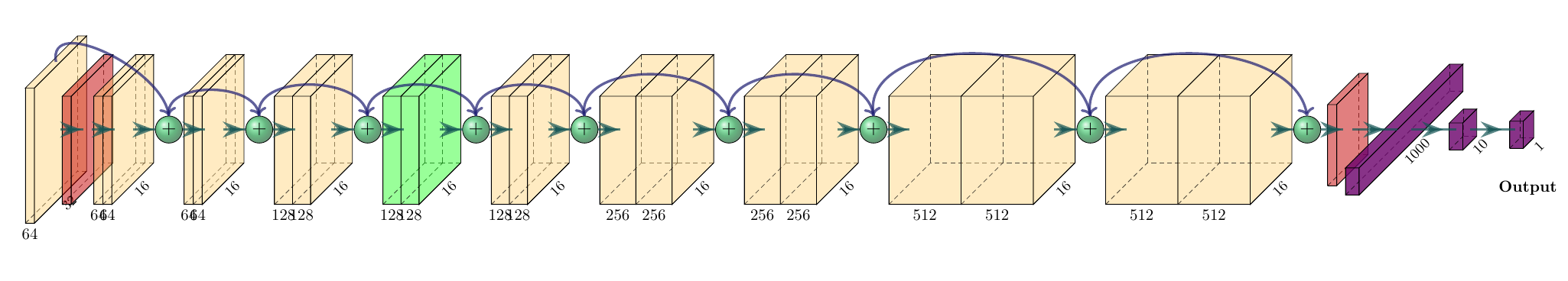}
        \label{fig:archis_CIF_Res18_02a}
    \end{subfigure}
    \begin{subfigure}{\textwidth}
        \centering
        \includegraphics[width=\linewidth]{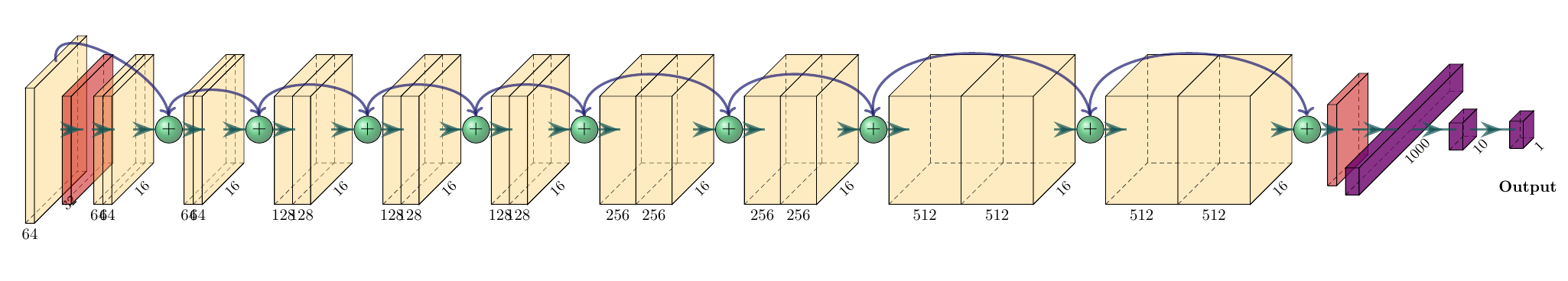}
        \label{fig:archis_CIF_Res18_02d}
    \end{subfigure}
    
    \begin{subfigure}{\textwidth}
        \centering
        \includegraphics[width=\linewidth]{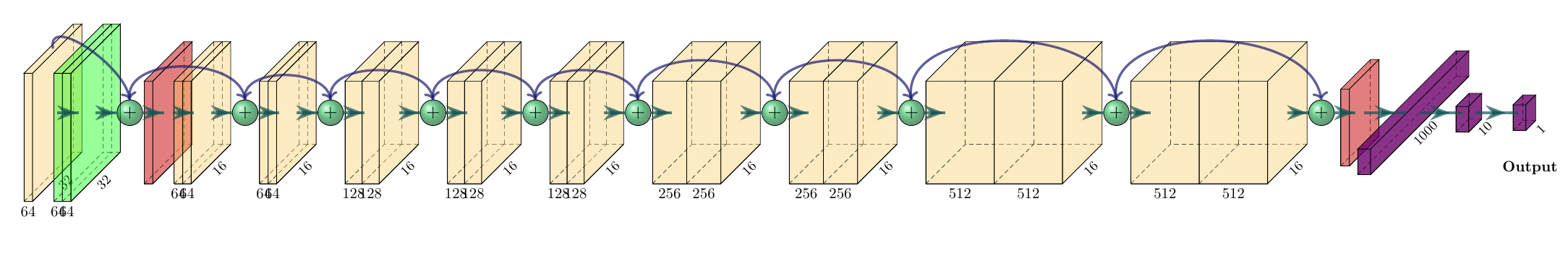}
        \label{fig:archis_CIF_Res18_03a}
    \end{subfigure}
    \begin{subfigure}{\textwidth}
        \centering
        \includegraphics[width=\linewidth]{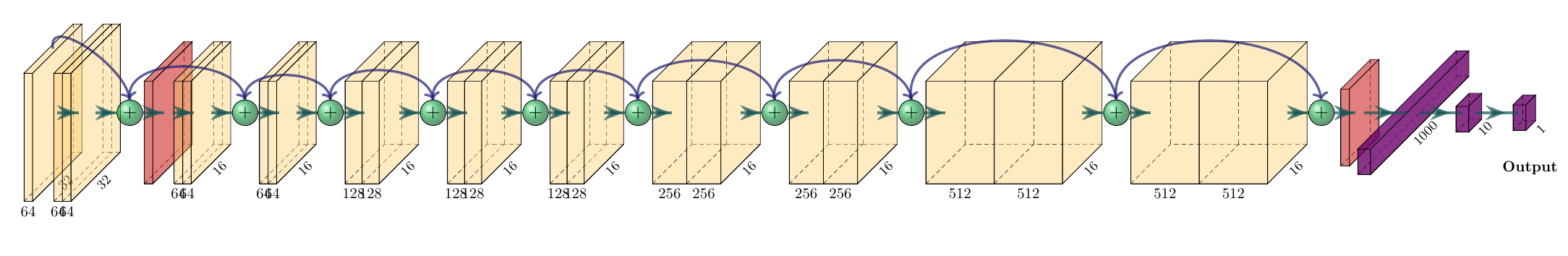}
        \label{fig:archis_CIF_Res18_03d}
    \end{subfigure}

    \begin{subfigure}{\textwidth}
        \centering
        \includegraphics[width=\linewidth]{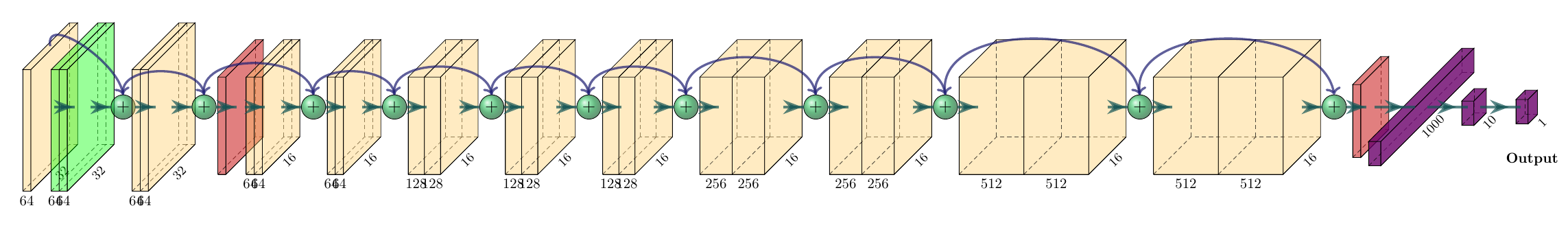}
        \label{fig:archis_CIF_Res18_04a}
    \end{subfigure}
    \begin{subfigure}{\textwidth}
        \centering
        \includegraphics[width=\linewidth]{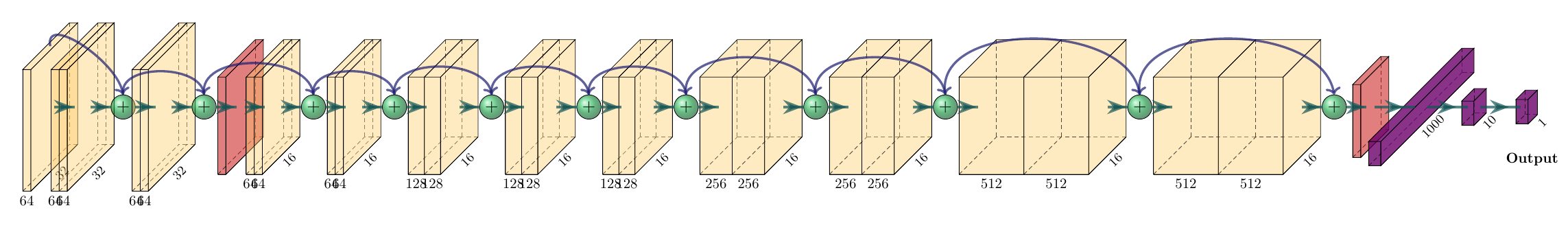}
        \label{fig:archis_CIF_Res18_04d}
    \end{subfigure}
    \begin{subfigure}{\textwidth}
        \centering
        \includegraphics[width=\linewidth]{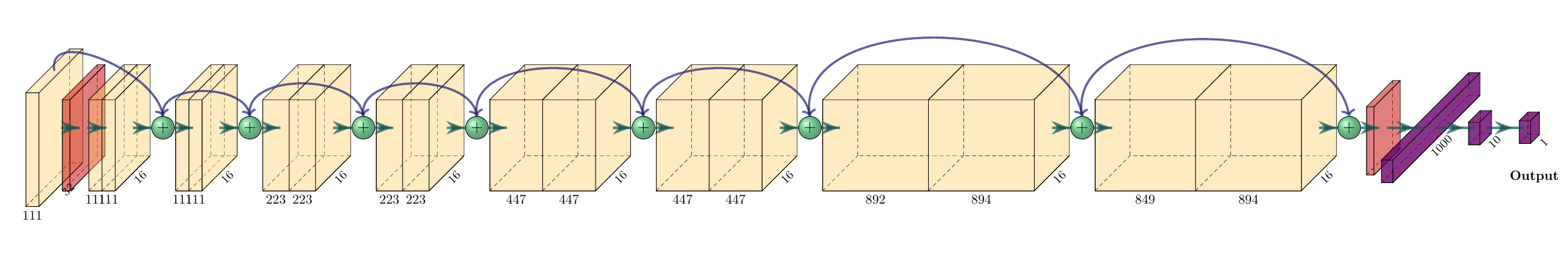}
        \label{fig:archis_CIF_Res18_04s}
    \end{subfigure}
\caption{Progress of network architecture on CIFAR10 with ResNet18 as initial architecture - first morphing routine.}
\label{fig:archis_CIF_Res18_1}
\end{figure*}

\begin{figure*}

    \begin{subfigure}{\textwidth}
        \centering
        \includegraphics[width=\linewidth]{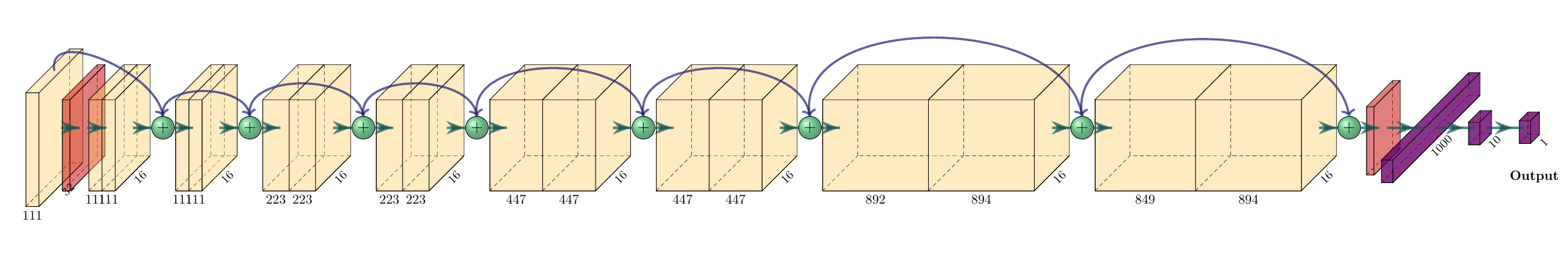}
        \label{fig:archis_CIF_Res18_05d}
    \end{subfigure}
    
    \begin{subfigure}{\textwidth}
        \centering
        \includegraphics[width=\linewidth]{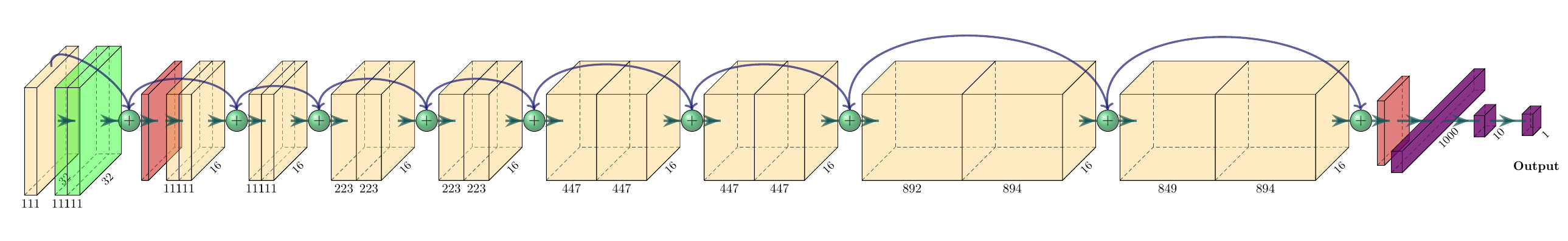}
        \label{fig:archis_CIF_Res18_06a}
    \end{subfigure}
    \begin{subfigure}{\textwidth}
        \centering
        \includegraphics[width=\linewidth]{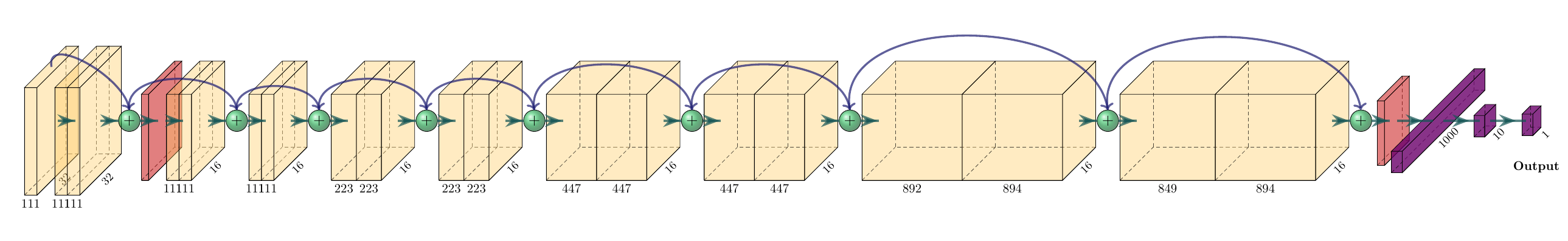}
        \label{fig:archis_CIF_Res18_06d}
    \end{subfigure}

    \begin{subfigure}{\textwidth}
        \centering
        \includegraphics[width=\linewidth]{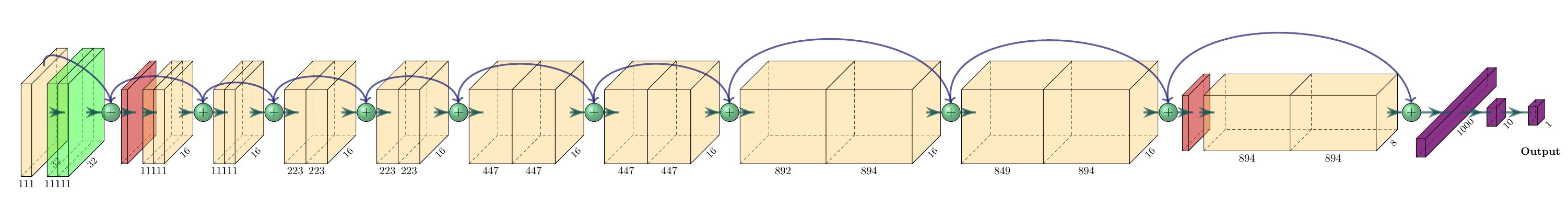}
        \label{fig:archis_CIF_Res18_07a}
    \end{subfigure}
    \begin{subfigure}{\textwidth}
        \centering
        \includegraphics[width=\linewidth]{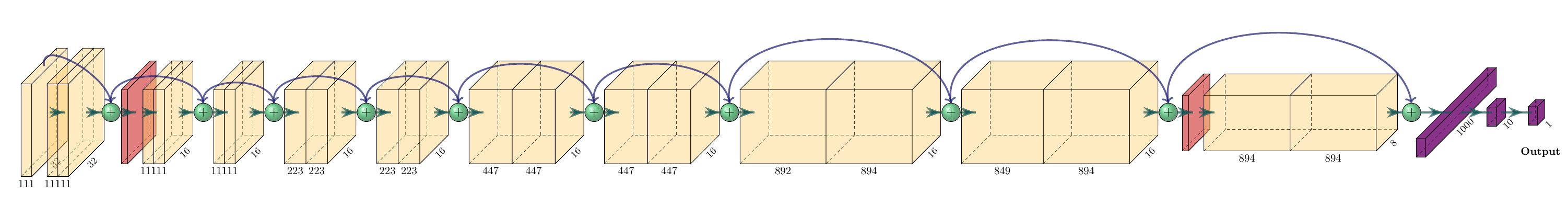}
        \label{fig:archis_CIF_Res18_07d}
    \end{subfigure}

    \begin{subfigure}{\textwidth}
        \centering
        \includegraphics[width=\linewidth]{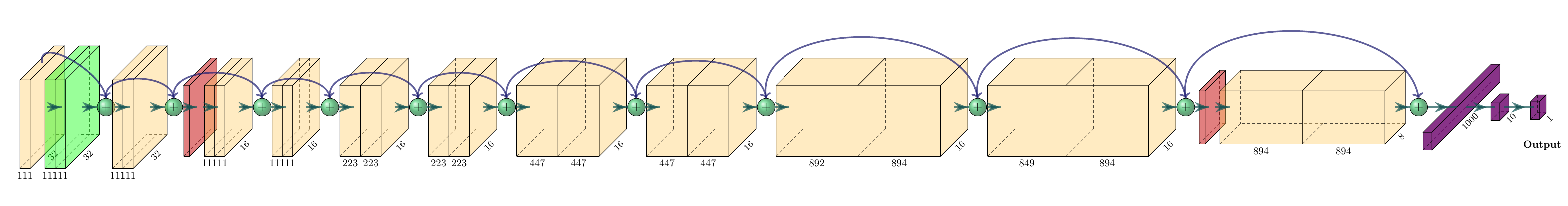}
        \label{fig:archis_CIF_Res18_08a}
    \end{subfigure}
    \begin{subfigure}{\textwidth}
        \centering
        \includegraphics[width=\linewidth]{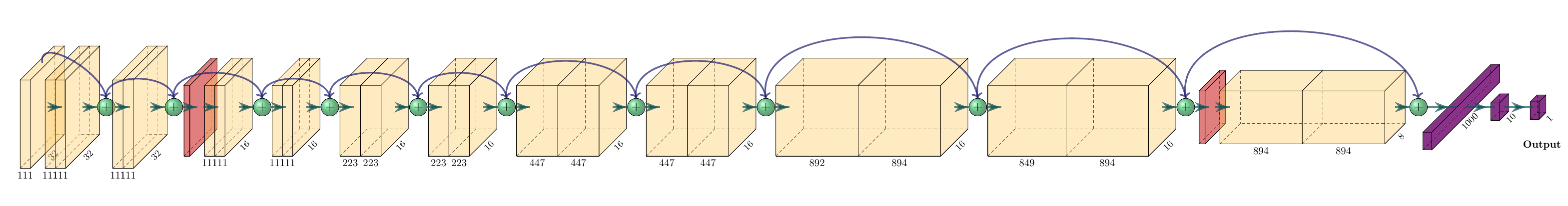}
        \label{fig:archis_CIF_Res18_08d}
    \end{subfigure}
    \begin{subfigure}{\textwidth}
        \centering
        \includegraphics[width=\linewidth]{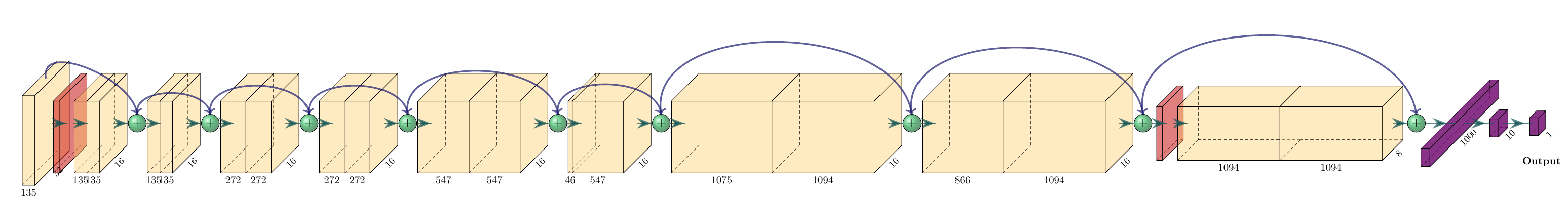}
        \label{fig:archis_CIF_Res18_08s}
    \end{subfigure}
    
\caption{Progress of network architecture on CIFAR10 with ResNet18 as initial architectur - second morphing routine}
\label{fig:archis_CIF_Res18_2}
\end{figure*}

\begin{figure*}

    \begin{subfigure}{\textwidth}
        \centering
        \includegraphics[width=\linewidth]{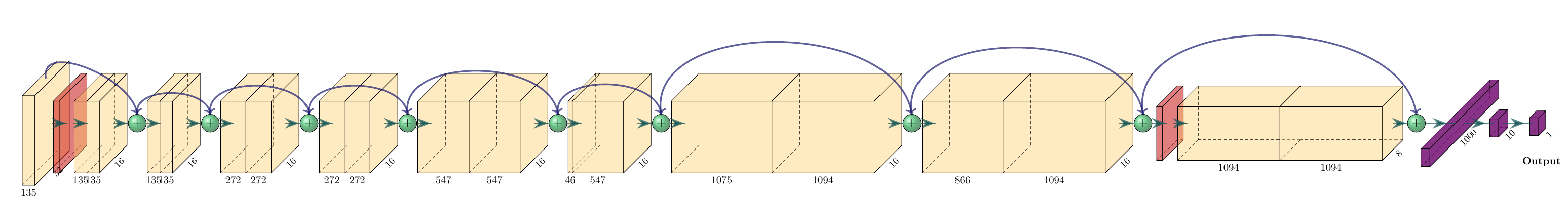}
        \label{fig:archis_CIF_Res18_09d}
    \end{subfigure}
    
    \begin{subfigure}{\textwidth}
        \centering
        \includegraphics[width=\linewidth]{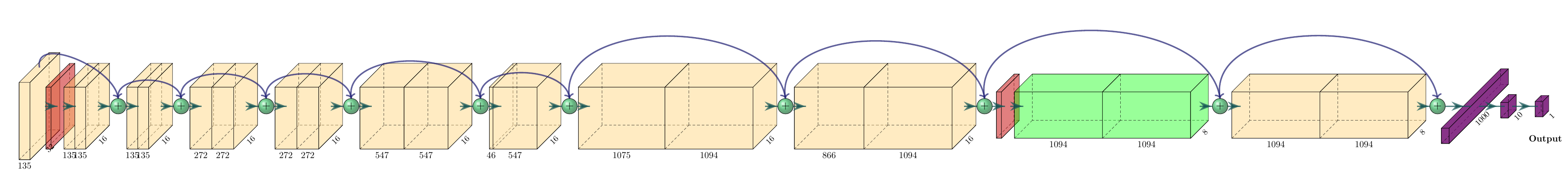}
        \label{fig:archis_CIF_Res18_10a}
    \end{subfigure}
    \begin{subfigure}{\textwidth}
        \centering
        \includegraphics[width=\linewidth]{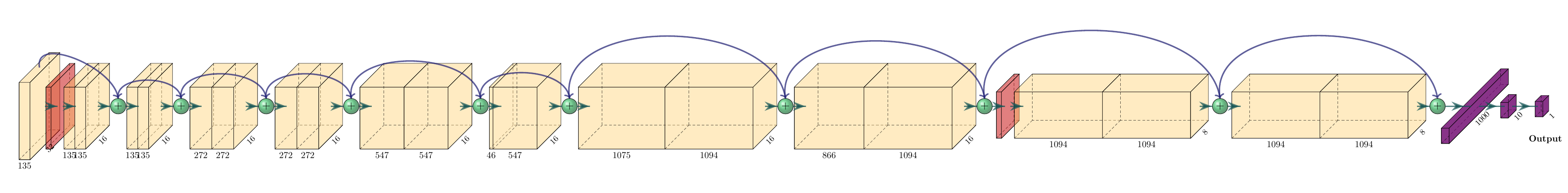}
        \label{fig:archis_CIF_Res18_10d}
    \end{subfigure}

    \begin{subfigure}{\textwidth}
        \centering
        \includegraphics[width=\linewidth]{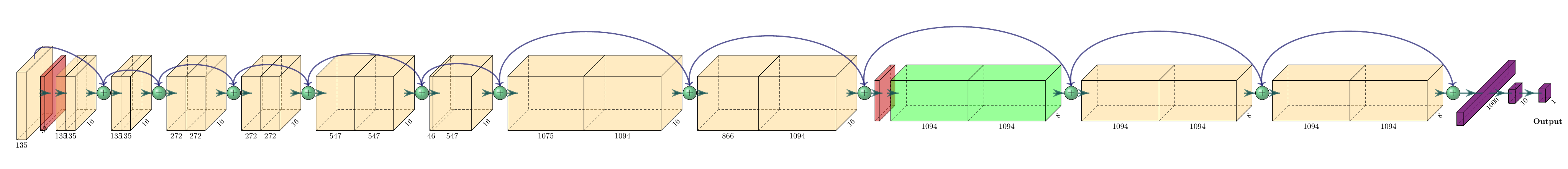}
        \label{fig:archis_CIF_Res18_11a}
    \end{subfigure}
    \begin{subfigure}{\textwidth}
        \centering
        \includegraphics[width=\linewidth]{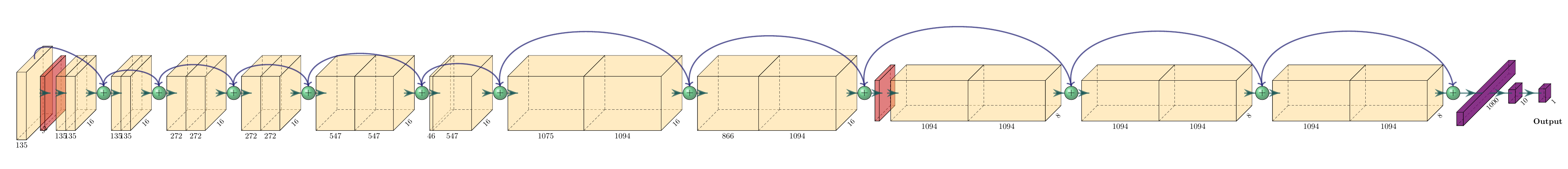}
        \label{fig:archis_CIF_Res18_11d}
    \end{subfigure}

    \begin{subfigure}{\textwidth}
        \centering
        \includegraphics[width=\linewidth]{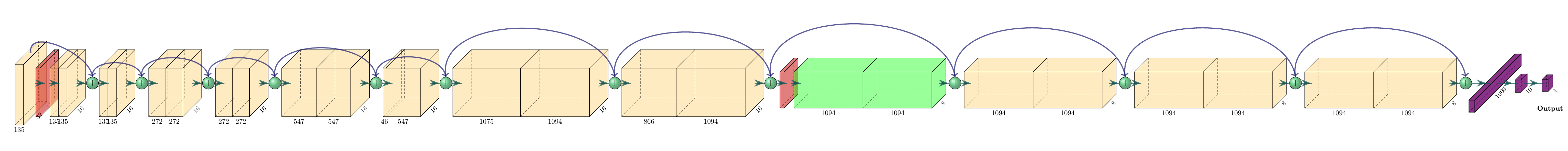}
        \label{fig:archis_CIF_Res18_12a}
    \end{subfigure}
    \begin{subfigure}{\textwidth}
        \centering
        \includegraphics[width=\linewidth]{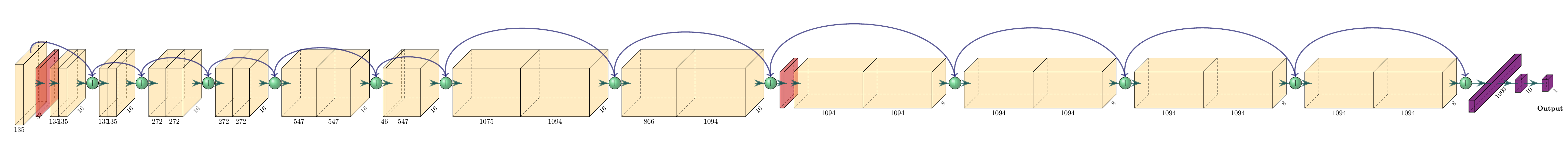}
        \label{fig:archis_CIF_Res18_12d}
    \end{subfigure}
    \begin{subfigure}{\textwidth}
        \centering
        \includegraphics[width=\linewidth]{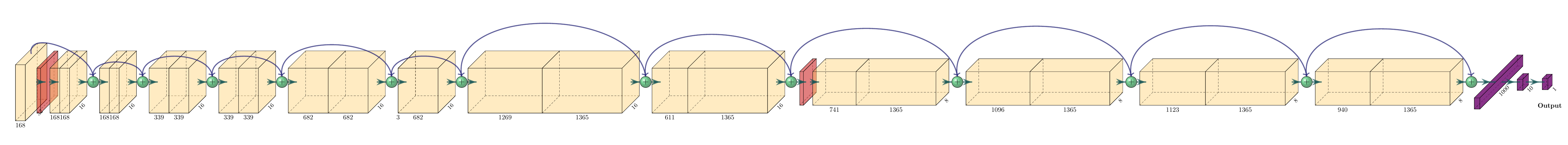}
        \label{fig:archis_CIF_Res18_12s}
    \end{subfigure}
    
\caption{Progress of network architecture on CIFAR10 with ResNet18 as initial architectur - third morphing routine}
\label{fig:archis_CIF_Res18_3}
\end{figure*}

In figures \ref{fig:archis_CIF_Res18_1}, \ref{fig:archis_CIF_Res18_2} and \ref{fig:archis_CIF_Res18_3}, every row shows one neural network architecture considered by the ResBuilder method, while rows with a green layerblock show one insertion step, the lines between a potential layerblock-removing step and the last row shows the concluding morphing routine. In insertion steps, there is always a new (green) block of two convolutional layers added to the current network architecture, while in the removing steps a block of layers is only removed from the architecture if the sum of the weights of at least one of the layers inside the block fell under the previous set threshold while training. The concluding morphing routine sets new widths of all convolutional layers according to \cite{Gordon17} and if there is a layer down to zero channels, its containing block is removed completely from the architecture.


\end{document}